\documentclass[10pt,twocolumn,letterpaper]{article}

\usepackage{colortbl}  
\usepackage{array}   
\usepackage{bbding}
\usepackage{amsthm,amsmath,amssymb}
\usepackage{mathrsfs}
\usepackage{multirow}
\usepackage{amssymb}
\usepackage{bbding}
\usepackage{algorithm}
\usepackage{algorithmic}

\usepackage{cvpr}              

\definecolor{cvprblue}{rgb}{0.21,0.49,0.74}
\usepackage[pagebackref,breaklinks,colorlinks,allcolors=cvprblue]{hyperref}

\def\paperID{4045} 
\def\confName{CVPR}
\def\confYear{2025}

\title{Divide and Conquer: Heterogeneous Noise Integration for \\ Diffusion-based Adversarial Purification}
\author{
Gaozheng Pei$^{1}$
Shaojie Lyu$^{2}$
Gong Chen$^{2}$
Ke Ma$^{1}$\thanks{Corresponding author}\ \ 
Qianqian Xu$^{3}$
Yingfei Sun$^{1}$
Qingming Huang$^{3,4,5*}$
\\
$^{1}$School of Electronic, Electrical and Communication Engineering, UCAS\quad
$^{2}$ Tencent Corporate \\ 
$^{3}$ Key Laboratory of Intelligent Information Processing, Institute of Computing Technology, CAS\\
$^{4}$School of Computer Science and Technology, UCAS\\
$^{5}$Key Laboratory of Big Data Mining and Knowledge Management, UCAS\\
\tt\small peigaozheng23@mails.ucas.ac.cn,\quad\{shaojielv,natchen\}@tencent.com\\
\tt\small xuqianqian@ict.ac.cn,\quad\{make,yfsun,qmhuang\}@ucas.ac.cn
}

\begin{document}
\maketitle
\begin{abstract}

\noindent Existing diffusion-based purification methods aim to disrupt adversarial perturbations by introducing a certain amount of noise through a forward diffusion process, followed by a reverse process to recover clean examples. However, this approach is fundamentally flawed: the uniform operation of the forward process across all pixels compromises normal pixels while attempting to combat adversarial perturbations, resulting in the target model producing incorrect predictions. Simply relying on low-intensity noise is insufficient for effective defense. To address this critical issue, we implement a heterogeneous purification strategy grounded in the interpretability of neural networks. Our method decisively applies higher-intensity noise to specific pixels that the target model focuses on while the remaining pixels are subjected to only low-intensity noise. This requirement motivates us to redesign the sampling process of the diffusion model, allowing for the effective removal of varying noise levels.
Furthermore, to evaluate our method against strong adaptative attack, our proposed method sharply reduces time cost and memory usage through a single-step resampling. The empirical evidence from extensive experiments across three datasets demonstrates that our method outperforms most current adversarial training and purification techniques by a substantial margin. Code is available at \url{https://github.com/GaozhengPei/Purification}.

\end{abstract}    
\section{Introduction}
\label{sec:intro}

\begin{figure}[ht]
    \centering
    \includegraphics[width=0.45\textwidth]{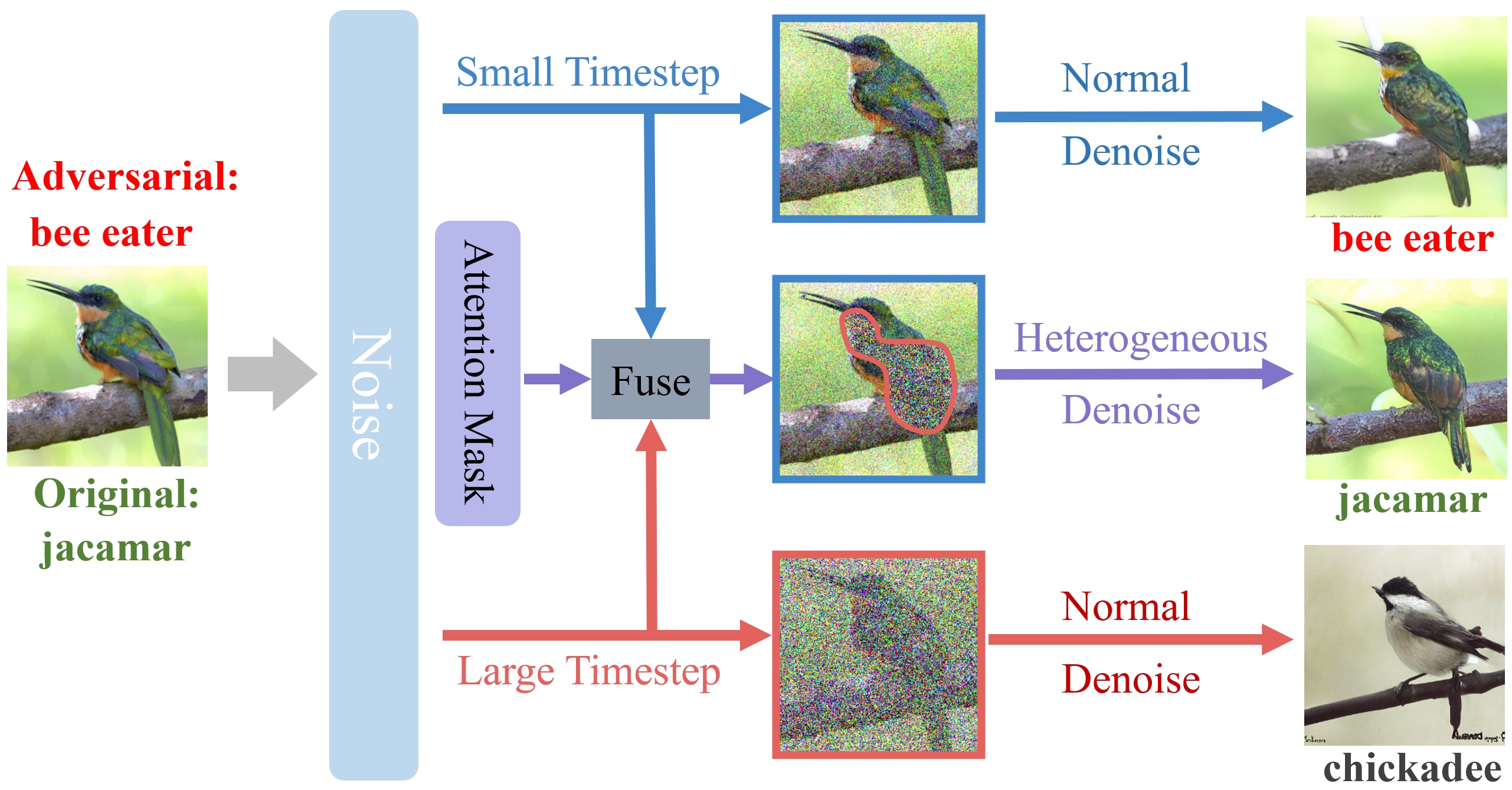}
    \caption{When the noise added to adversarial samples is minimal (top), the adversarial perturbations remain intact and cannot be removed. On the other hand, if the noise intensity is excessive (bottom), it can distort the semantic information. Our approach employs an attention mask to introduce varying intensities of noise across different areas (middle). This technique effectively strikes a balance between preserving semantic information and mitigating adversarial perturbations.}
    \label{others&ours}
\end{figure}


Adversarial perturbations \cite{adversarial}, which are often imperceptible to humans, can significantly alter the predictions of neural networks. To address these risks \cite{pei2024exploring,liang2020efficient,liang2022parallel,liang2022large}, various defense mechanisms have been proposed. One common approach is adversarial training \cite{freeadv,fastadv,jia2024improving}, which involves generating adversarial samples according to a specific attack method before the test phase and then re-training the model to improve its robustness. However, this approach often struggles to defend against unknown attacks. Another technique, known as adversarial purification \cite{diffpure,GDMAP,mimic,meng2017magnet,hwang2023aid,shi2021online}, seeks to remove perturbations from adversarial examples during the test phase. This approach does not rely on any assumptions about the attack methods and eliminates the overhead of repetitive training. Its flexibility and strong generalization have drawn increasing interest. In recent years, diffusion models \cite{DDPM,ImprovedDDPM} have garnered attention due to their impressive denoising abilities and capacity to generate high-quality images, making them a promising option for adversarial purification.

Existing diffusion-based adversarial purification methods \cite{diffpure,mimic,GDMAP,Contrastive} follow the traditional forward process which attempts to add uniform noise across different regions of the images and then recover the clean images through reverse process. From Figure \ref{others&ours} we can observe that such a manner can not balance eliminating perturbations and preserving semantic information. If the added noise is intense, it will remove the adversarial perturbations but damage the semantic information, causing the wrong prediction of both natural and adversarial examples. In contrast, if the noise intensity is weak, it may not effectively purify the adversarial perturbations, resulting in a successful evasion attack. Therefore, how to achieve a good balance between the elimination of adversarial perturbations and the preservation of semantic information remains challenging.

In this paper, we propose a novel diffusion-based adversarial purification method from the perspective of neural network interpretability. During making decisions, the neural network spontaneously allocates varying levels of attention to different regions of the image. This phenomenon inspired us to use attention maps as a prior guided diffusion model to impose heterogeneous noise on different regions of an image as forward process. By eliminating adversarial perturbations in the critical areas through high-intensity noise, we can substantially mitigate their effect on the prediction of the target model. Conversely, for regions that receive less attention, we introduce low-intensity noise to preserve as much semantic information as possible in the original image. To recover a clean image from one containing heterogeneous noise, we design a two-stage denoising process in accordance with the attention map as a reverse process of the diffusion model. Specifically, if the sampling timestep falls between two different noise levels, we apply inpainting techniques to remove localized high-intensity noise. If the current timestep is below that of the smaller noise level, we use the standard denoising process to remove global weak noise. Additionally, by replacing multi-step resampling with an improved single-step resampling approach, this strategy results in a 90\% reduction in both time consumption and GPU memory usage enables evaluate our method against strong adaptive attacks. Overall, our contributions can be summarized as follows:
\begin{itemize}
    \item We propose a novel forward process that adds noise of varying intensities based on the classifier's attention to different regions, achieving a balance between preserving semantic information and eliminating adversarial perturbations.
    \item To recover the clean samples, we design a two-stage denoising process: at a larger timestep, we convert it into an inpainting problem to remove localized strong noise, and at a smaller timestep, we apply a standard sampling process to eliminate global weak noise.
    \item To evaluate our method against strong adaptive attacks. We replace the multi-step resampling with an improved single-step resampling, this not only achieves harmonious image generation but also significantly reduces time cost and memory usage.
    \item Extensive experiments on three datasets show that our method outperforms other methods by a promising improvement against various adversarial attacks.
\end{itemize}

\section{Related Work}
Adversarial purification aims to remove adversarial perturbations from samples at test phase before feeding them into classifier model. 
Adversarial purification methods can be broadly divided into two categories: training-based methods, which require training on the dataset, and diffusion based methods which doesn't require training and is not necessary to access the training dataset.
\subsection{Training-Based Adversarial Purification}
\cite{pixeldefend} shows empirically that adversarial examples mainly lie in the low probability regions of the training distribution and move them back towards the training distribution. \cite{defensegan} first models the distribution of clean samples and then attempts to find the closest sample within the clean sample distribution to the adversarial example during inference. \cite{SSP} generates perturbed images using Self-supervised Perturbation (SSP) attack that disrupts the deep perceptual features and projects back the perturbed images close to the perceptual space of clean images. \cite{Invariant} propose to learn generalizable invariant features across attacks via encoder by playing zero-sum game and reconstruct the original image via decoder. \cite{ATOP} disrupt dversarial perturbations structure through random transform and reconstruct the image using a purifier model. \cite{RFGSM} experimentally show that FGSM-AT networks overfit to unknown adversarial examples attacked again by FGSM. In the test phase, adversarial examples are fed to classifier after performing FGSM "purification". However, these methods require training using the training dataset, which is very time-consuming and lacks generalizability.
\subsection{Diffusion-Based Adversarial Purification}
\cite{APSGM} show that an EBM trained with Denoising Score-Matching can quickly purify attacked images within a few steps. \cite{diffpure} first diffuse samples with a small amount of noise and then recover the clean image through a reverse generative process. The paper also theoretically proves that the time step cannot be too large or too small. \cite{GDMAP} proposes to use the adversarial image as guidance in the reverse process which encourages the purified image is consistent with the original clean image. \cite{densepure} gets multiple reversed samples via multiple runs of denoising. Then the samples are passed into the classifier, where the final prediction is obtained through classifier voting. \cite{Contrastive} design the forward process with the proper amount of Gaussian noise added and the reverse process with the gradient of contrastive loss as the guidance of diffusion models for adversarial purification. However, adding noise of the same intensity to different regions of the images, these methods is difficult to balance preserving semantic information and eliminating adversarial perturbations.

 \begin{figure*}[!t]
    \centering
    \includegraphics[width=1.0\textwidth]{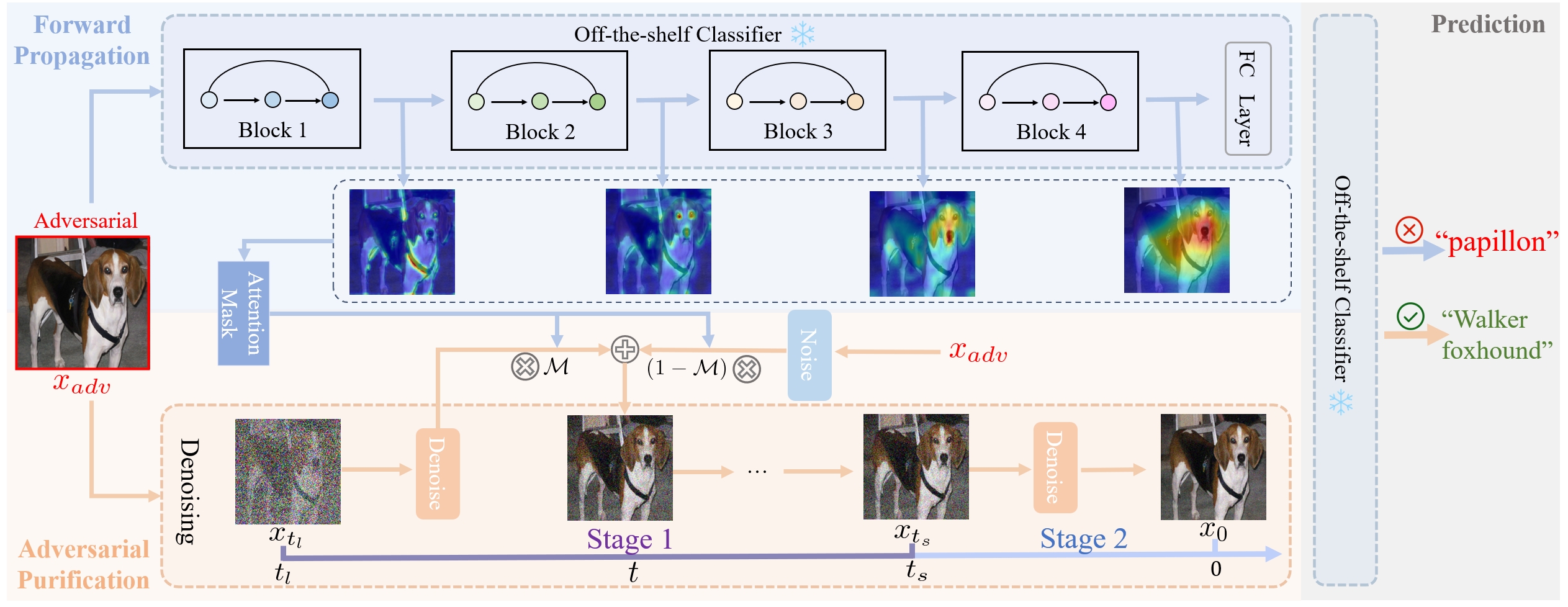}
    \caption{Pipeline of our method. Given an adversarial image, we extract the attention maps of each block during the forward propagation of the classifier, and construct an attention mask $\mathcal{M}$. We then execute the denoising process of the diffusion model we designed. It can be seen that the samples purified by our method can be correctly classified.}
    \label{method}
\end{figure*}
\section{Methodology}
Given an example $(\boldsymbol{x}, \boldsymbol{y})$, an adversary can create a small perturbation to generate an adversarial example $\boldsymbol{x}_{adv}$ such that the classifier $\boldsymbol{f}$ makes an incorrect prediction: $\boldsymbol{y} \neq \boldsymbol{f}(\boldsymbol{x}_{adv})$. Existing diffusion-based purification methods obtain a noisy image $\boldsymbol{x}(t)$ at timestep $t$ through the forward process and then recover the clean (or purified) image $\hat{\boldsymbol{x}}$ through the reverse process. And we hope $\boldsymbol{f}(\hat{\boldsymbol{x}}) = \boldsymbol{f}(\boldsymbol{x})$. From the theoretical analysis of \cite{diffpure}, the timestep $t$ should satisfy: 1) $t$ should be large enough to eliminate perturbations. 2) $t$ should be small enough that $\hat{\boldsymbol{x}}$ still maintains semantic consistency with $\boldsymbol{x}$. We find that these two objectives are contradictory. 
\\Motivated by the interpretability of neural networks, the attention map 
is primarily used to encode the input spatial regions that the network focuses on most when making output decisions, such as in image classification. This suggests that, for an adversarial example, adversarial perturbations in the regions where the classifier focuses most significantly contribute to causing the target classifier to make incorrect predictions. The attention maps for each layer of the target classifier can be obtained with a single forward propagation
Therefore, we propose introducing high-intensity noise to eliminate adversarial perturbations of the regions with more attention value and applying low-intensity noise to the remaining area to preserve as much semantic information as possible.

\subsection{Heterogeneous Forward Process}
\label{forward}
For the adversarial example \(\boldsymbol{x}_{adv}\) at timestep \(t = 0\), i.e., \(\boldsymbol{x}(0) = \boldsymbol{x}_{adv}\), we can obtain the noisy images at timestep \(t_l\) and \(t_s\) with $t_l > t_s$ as follows:

\begin{subequations}
    \label{eq:forward}
    \begin{align}
        \boldsymbol{x}(t_l) &= \sqrt{\overline{\alpha}(t_l)}\cdot \boldsymbol{x}_{adv} + \sqrt{1 - \overline{\alpha}(t_l)}\cdot \boldsymbol{\epsilon}\label{x(t_l)},\\
        \boldsymbol{x}(t_s) &= \sqrt{\overline{\alpha}(t_s)}\cdot \boldsymbol{x}_{adv} + \sqrt{1 - \overline{\alpha}(t_s)}\cdot \boldsymbol{\epsilon}\label{x(t_s)},
\end{align}
\end{subequations}
where $\{\sqrt{\overline{\alpha}(t)}\}$ is a series of given parameters and $\boldsymbol{\epsilon} \sim \mathcal{N}(0, \boldsymbol{I})$.
The clean (or purified) images \(\boldsymbol{\hat{x}}_{t_l}\) and \(\boldsymbol{\hat{x}}_{t_s}\) can be obtained through the normal reverse process \cite{diffpure,DDPM}. For the smaller timestep \(t_s\), the recovered image \(\boldsymbol{\hat{x}}_{t_s}\) retains certain adversarial perturbation, resulting in \(\boldsymbol{f}(\boldsymbol{x}_{adv}) = \boldsymbol{f}(\boldsymbol{\hat{x}}_{t_s}) \neq \boldsymbol{f}(\boldsymbol{x})\). Conversely, at the larger timestep \(t_l\), the clean image \(\boldsymbol{\hat{x}}_{t_l}\) loses critical semantic information, leading to \(\boldsymbol{f}(\boldsymbol{x}_{adv}) \neq \boldsymbol{f}(\boldsymbol{\hat{x}}_{t_l}) \neq \boldsymbol{f}(\boldsymbol{x})\). To effectively aggregate the differing impacts on various regions of the images at distinct timesteps, we construct an attention mask to integrate \eqref{x(t_l)} and \eqref{x(t_s)}.

Consider a neural network that consists of $M$ blocks (\textit{e.g.} ResNet-50 \cite{resnet} consists of $4$ blocks). For $m$-th block, we extract the output of the last activation function (\textit{e.g.} Relu \cite{relu} or Swish \cite{swish}) as follows: 
\begin{equation}
    \boldsymbol{G}_m  = \left( \sum_{i=1}^{C_m} |\boldsymbol{A}_{mi}|^p \right)^{\frac{1}{p}}\in \mathbb{R}^{H_m \times W_m}, m\in[M],
\end{equation}
where $p\in\mathbb{N}$, $C_m, H_m, W_m$ are the dimension of channels, height, and width respectively, $\boldsymbol{A}_m$ denotes the activation of \( m \)-th block and $\boldsymbol{A}_{mi}$ denotes the $i$-th slice of $\boldsymbol{A}_m$ along the channel dimension. Then we apply interpolation techniques to ensure it matches the original image size. By upsampling $\boldsymbol{G}_m$ with bilinear function $\text{Bi}(\cdot)$, we obtain attention map of $m$-th block:
\begin{equation}
    \text{AM}_m = \phi(\text{Bi}(\boldsymbol{G}_m)),
\end{equation}
where $\phi(\cdot)$ is the spatial softmax operation. Moreover, we build the attention mask with all attention maps $\{\text{AM}_m\}$:
\begin{equation}
    \label{attention-mask}
    \mathcal{M} = \bigcup_{m=1}^M \mathbb{I}\ [\text{AM}_m>\tau]
\end{equation}
where $\mathbb{I}\ [\cdot]$ is the Iverson bracket and $\tau$ is the threshold to distinguish important and ordinary regions. With the attention mask $\mathcal{M}$ and two noisy images from Eq.\eqref{eq:forward}, $\boldsymbol{x}(t_s),\boldsymbol{x}(t_l), t_s<t_l$, the result of the proposed heterogeneous forward process is 
\begin{equation}
    \label{x_a&b}
    \boldsymbol{x}(t_l,t_s) = \boldsymbol{x}(t_l) \odot \mathcal{M} + \boldsymbol{x}(t_s) \odot (1-\mathcal{M}),
\end{equation}
where $\odot$ is the Hadamard product. 
\subsection{Heterogeneous Denoising Process}
\label{Denoising}

\begin{figure}[!t]
    \centering
    \includegraphics[width=0.5\textwidth]{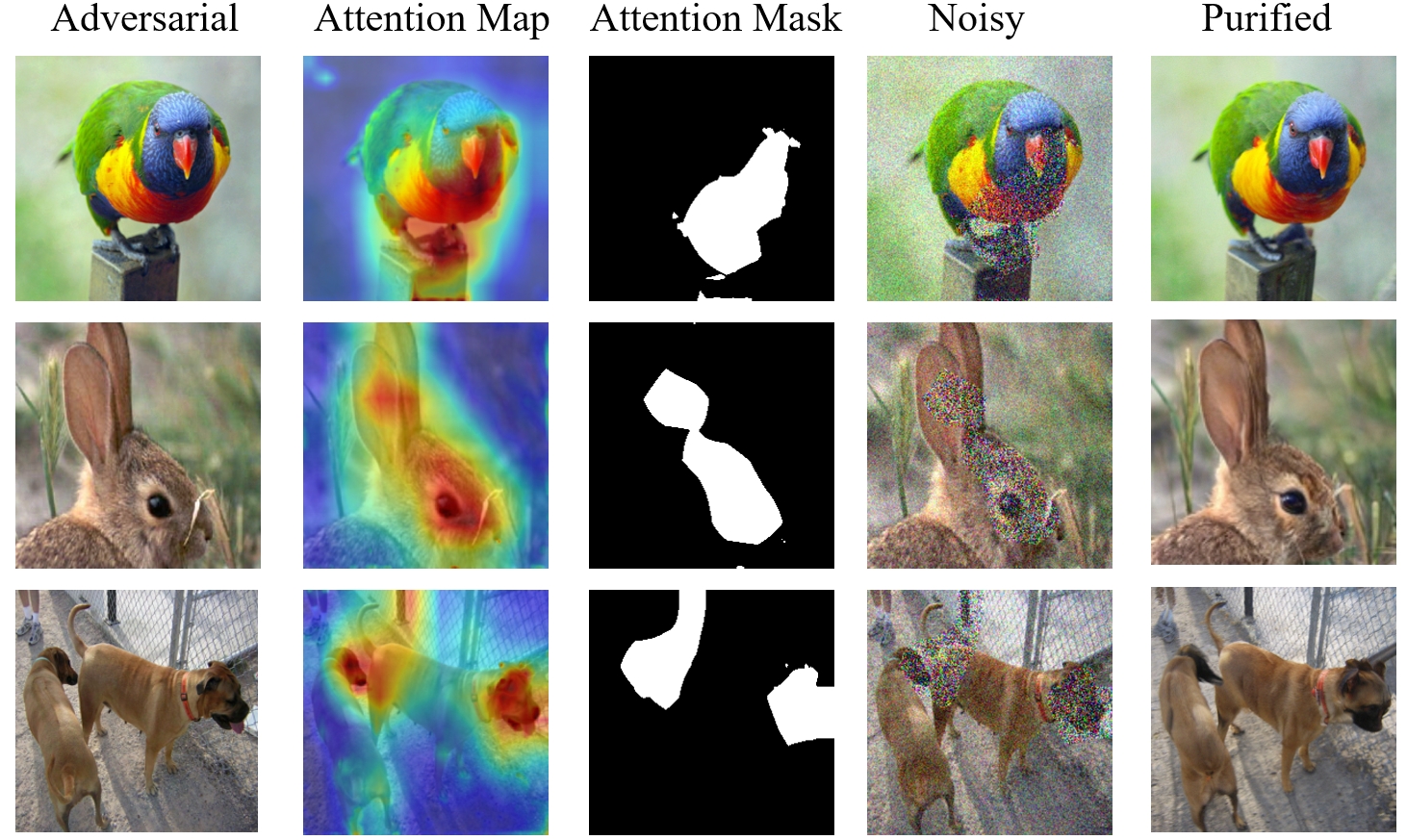}
    \caption{Visualization of attenion map, attention mask and images purified by our method.}
    \label{visual}
\end{figure}

To recover the clean image from \eqref{x_a&b}, we design a two-stage denoising process. For the current timestep \(t\), we consider the following two cases:\\
\textbf{Stage 1 ($\boldsymbol{t_s<t<t_l}$)}\quad When the current timestep $t$ lies between $t_s$ and $t_l$. For the pixels inside the attention mask $\mathcal{M}$, they should be predicted from the previous timestep $t+1$. But for the pixels outside the attention mask $\mathcal{M}$, they can be obtained by further adding noise forward from \eqref{x_a&b}. Therefore, this case can be viewed as an inpainting problem \cite{repaint}.  
For the image regions outside the attention mask $\mathcal{M}$, since only the forward process is involved. Therefore, we can get the pixel value of these regions at timestep $t$ equivalently as follows:
\begin{equation}
\label{known}
    \boldsymbol{x}(t)^{known} = \sqrt{\overline{\alpha}(t)} \boldsymbol{x}_{adv} + \sqrt{1 - \overline{\alpha}(t)}\boldsymbol{ \epsilon}.
\end{equation}
 For the image regions within the attention mask, the pixel value of these regions at timestep $t$ should be predicted by the diffusion model:
\begin{equation}
    \label{unknown}
    \boldsymbol{x}(t)^{unknown}
    \sim\mathcal{N}(\mu_\theta(\boldsymbol{x}(t+1), t+1), \Sigma_\theta(\boldsymbol{x}(t+1), t+1))
\end{equation}
Where $\theta$ is the parameters of the diffusion model. $\mu_\theta$ and $\Sigma_\theta$ are mean and variance, respectively. Then we combine $x(t)^{known}$ and $x(t)^{unknown}$ together based on the attention mask $\mathcal{M}$ to get the  $\boldsymbol{x}(t)$ at timestep $t$:
\begin{equation}
\label{merge}
\boldsymbol{x}(t) = \mathcal{M} \odot \boldsymbol{x}(t)^{known} + (1 - \mathcal{M}) \odot \boldsymbol{x}(t)^{unknown},
\end{equation}
where $\odot$ is Hadamard product. We can remove the localized strong noise through the Stage 1.\\
\textbf{Stage 2 ($\boldsymbol{t<t_s}$)}\quad When the current timestep $t$ is less than $t_s$, since the noise intensity is the same both inside and outside the attention mask. Therefore, the attention mask $\mathcal{M}$ is no longer needed. At this point, we can proceed with the standard sampling process of the diffusion model. We can get $\boldsymbol{x}(t)$ predicted from the previous timestep $\boldsymbol{x}(t+1)$ as follows:
\begin{equation}
    \label{reverse}
\boldsymbol{x}(t)
\sim\mathcal{N}(\mu_\theta(\boldsymbol{x}(t+1), t+1), \Sigma_\theta(\boldsymbol{x}(t+1), t+1))
\end{equation}
Following Stage 2, we will eliminate global weak noise and get the clean image. The complete algorithm process can be found in the Algorithm \ref{sampling-argotithm}. Additionally, there are some visualizations of our method in the Figure \ref{visual}.

\begin{algorithm}[htbp]
\caption{Heterogeneous Denoising Process}
\label{sampling-argotithm}
\begin{algorithmic}
\REQUIRE Sample $x_0$, Attention Mask $\mathcal{M}$,  timestep $t_s$ and $t_l$, Diffusion model parameters $\theta$.


\FOR{$t=t_l,...,1$}

\STATE\textbf{if} $t_s \leq t \leq t_l$ \textbf{then} 
\COMMENT{Stage 1}
\STATE\quad $\epsilon \sim \mathcal{N}(0, \mathbf{I}) \text{ if } t > 1, \text{ else } \epsilon = 0$
\STATE\quad$x_{t-1}^{known} = \sqrt{\overline{\alpha}_{t-1}} x_0 + \sqrt{1-\overline{\alpha}_{t-1}}\epsilon$
\STATE\quad $z \sim \mathcal{N}(0, \mathbf{I}) \text{ if }  t > 1, \text{ else } z = 0$
\STATE\quad $x_{t-1}^{unknown} = \frac{1}{\sqrt{\alpha_t}} \left( x_t - \frac{\beta_t}{\sqrt{1-\bar{\alpha}_t}} \epsilon_\theta(x_t, t) \right) + \sigma_t z$
\STATE\quad$x_{t-1} = \mathcal{M} \odot x_{t-1}^{known} + (1 - \mathcal{M}) \odot x_{t-1}^{unknown}$
\STATE \textbf{else if} $t<t_s$ \textbf{then} \COMMENT{Stage 2} 

\STATE\quad $z \sim \mathcal{N}(0, \mathbf{I}) \text{ if }  t > 1, \text{ else } z = 0$ 
\STATE\quad $x_{t-1} = \frac{1}{\sqrt{\alpha_t}} \left( x_t - \frac{\beta_t}{\sqrt{1-\bar{\alpha}_t}} \epsilon_\theta(x_t, t) \right) + \sigma_t z$ 

\STATE\textbf{for} $u=1,2,...U$ \textbf{do} \COMMENT{Resampling}
\STATE\quad $z \sim \mathcal{N}(0, \mathbf{I}) \text{ if }  t > 1, \text{ else } z = 0$
\STATE\quad$x_{t-1+u} = \sqrt{\alpha_{t-1+u}} x_{t+u-2} + \sqrt{1 - \alpha_{t-1+u}} z$
\STATE\textbf{end for}

\STATE get $x_{t-1}$ using \eqref{ddim} 
\ENDFOR

\end{algorithmic}
\end{algorithm}

\section{Adaptive Attack to Diffusion Purification }
\label{Time and Space Efficiency} 
\begin{figure}[!t]
    \centering
\includegraphics[width=0.45\textwidth]{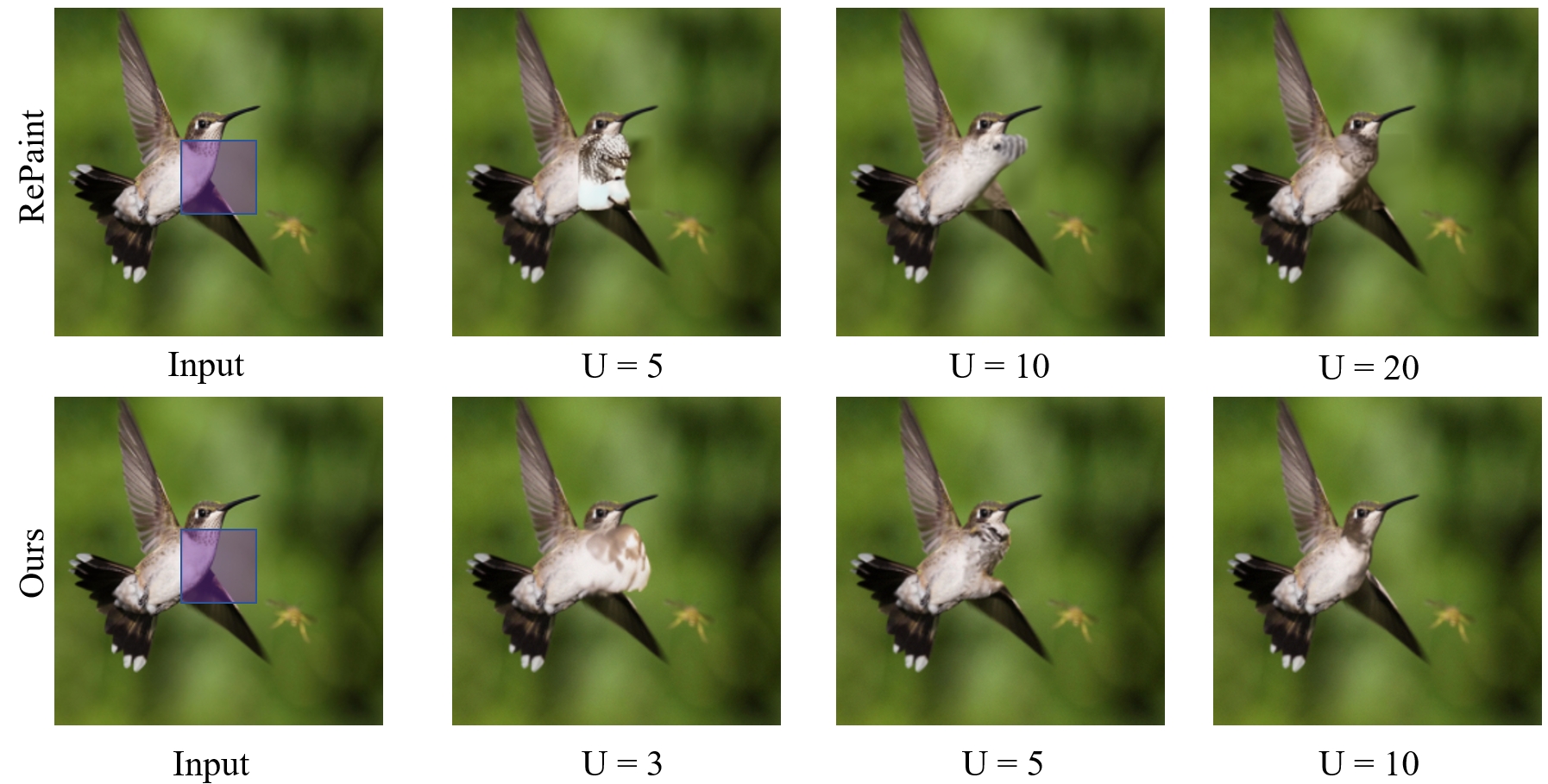}
    \caption{Our method maintains semantic consistency between the masked areas and the surrounding pixel values when $U=10$. Note that no matter what $U$ is, our method requires only one additional call to the denoising network. }
    \label{efficiency}
\end{figure}

When directly applying the method described in \ref{Denoising}, we observe the same phenomenon as \cite{repaint} that although the region recovered in Stage 2 matches the texture of the neighboring region, it is semantically incorrect.
We can observe this phenomenon in the Figure \ref{efficiency}. \cite{repaint} addresses this issue via muti-step resampling by diffusing the output $\boldsymbol{x}(t)$ back to $\boldsymbol{x}(t+1)$: 
\begin{equation} 
\label{back}
\boldsymbol{x}(t+1) = \sqrt{\alpha(t)} \boldsymbol{x}(t) + \sqrt{1 - \alpha(t)} \boldsymbol{\epsilon},
\end{equation}
and it is necessary to fully execute \ref{Denoising} again to get $\boldsymbol{x}(t)$. From Figure \ref{efficiency} we can find that this process needs to be repeated $U=20$ times to ensure that the masked region becomes semantically consistent with the surrounding pixels. To evaluate our method against strong adaptive attacks \cite{diffpure}, it is necessary to compute the full gradient of the whole sampling process to construct adversarial examples. To reduce the time cost and memory usage, we optimized the multi-step resampling into a single resampling process which achieves the same functionality.
\\The intuition behind our improvement is that repeating \eqref{back} and \ref{Denoising} for $U$ times can be replaced by
 that first diffuse $\boldsymbol{x}(t)$ to $\boldsymbol{x}(t+U)$ and then denoise it back to $\boldsymbol{x}(t)$ from $\boldsymbol{x}(t+U)$ in a single step using DDIM \cite{ddim} as follows:
\begin{equation}
\label{ddim}
\begin{aligned}
    \boldsymbol{x}_{t-1} &= \sqrt{\alpha_{t-1}} \left( \frac{\boldsymbol{x}_{t-1+U} - \sqrt{1 - \alpha_{t-1+U}} \epsilon_{\theta}(\boldsymbol{x}_{t-1+U})}{\sqrt{\alpha_{t-1+U}}} \right) \\
    &+ \sqrt{1 - \alpha_{t-1} - \sigma^2_{t-1+U}} \cdot \epsilon_{\theta}(\boldsymbol{x}_{t-1+U}) + \sigma_{t-1+U} \boldsymbol{\epsilon}.
\end{aligned}
\end{equation}
From the bottom panel of Figure \ref{efficiency}, we find that our method can also achieve semantically consistency when $U=10$. This means that at each timestep, we only need to invoke the denoising network an additional time. Therefore, the GPU memory needs to store the  $2 * T$ complete computation graphs of the denoising network which enables the computation of the gradient of our method on consumer-grade GPU with only 24G memory. Also, the time cost is greatly reduced. The analysis of time cost and memory usage can refer to the Appendix.

\section{Experiment}
\subsection{Experimental Settings}
\textbf{Datasets and network architectures.}\quad Three datasets is used to evaluate including CIFAR-10, SVHN, and ImageNet \cite{imagenet}. We compare our results with several leading defense methods as listed in the standardized benchmark RobustBench \cite{robustbench} for CIFAR-10 and ImageNet, and also test different adversarial purification methods. We use two common classifier architectures: WideResNet-28-10 and WideResNet-70-16 \cite{wide} for CIFAR-10. For SVHN,  WideResNet-28-10 serves as the backbone, while for ImageNet, ResNet-50 \cite{resnet} is used as the backbone. \\
\textbf{Adversarial attack methods.}\quad Strong adaptive attack is tested for our method and other adversarial purification methods. We use the commonly known AutoAttack \cite{autoattack} under $\ell_\infty$ and $\ell_2$ threat models. Additionally, the projected gradient descent (PGD) attack \cite{PGD} is evaluated on our method, as recommended in \cite{robust-evaluation}. Due to the randomness introduced by the diffusion and denoising processes, Expectation Over Time (EOT) \cite{athalye2018obfuscated} is adapted for these adaptive attacks. Moreover, we apply the BPDA+EOT \cite{hill2021stochastic} attack to ensure a fair comparison with other adversarial purification methods. Finally, following the  suggestion of \cite{robust-evaluation}, a surrogate process is used to obtain the gradient of the reverse process for white-box attacks.\\
\textbf{Pre-trained Models.} \quad
We use the unconditional CIFAR-10 checkpoint of EDM provided by NVIDIA \cite{chekpoint-cifar10} on CIFAR-10. 256x256 diffusion (unconditional) checkpoint from guided-diffusion library is adopted for the ImageNet experiments. For CIFAR-10, the pre-trained classifier is downloaded from RobustBench \cite{robustbench}, while for ImageNet, the weight of the classifier is provided by PyTorch libray.  \\
\textbf{Evaluation metrics.} \quad To assess the performance of defense methods, we use two metrics: standard accuracy, measured on clean samples, and robust accuracy, measured on adversarial examples. Due to the high computational cost of evaluating models against adaptive attacks, a fixed subset of 512 images are randomly sampled from the test set for robust evaluation, which is the same as  \cite{diffpure,robust-evaluation}. In all experiments, the mean and standard deviation over three runs are reported to assess both standard and robust accuracy.\\
\textbf{Implementation details.} \quad We follow the settings described in \cite{robust-evaluation}. We evaluate diffusion-based purification methods using the PGD+EOT attack with 200 update iterations, and BPDA and AutoAttack with 100 update iterations, except for ImageNet, which uses 20 iterations. We set the number of EOT to 20. The step size is 0.007. For randomized defenses, such as \cite{diffpure,robust-evaluation,mimic,Contrastive}, we use the random version of AutoAttack, and for static defenses, we use the standard version. 
\begin{table}[htbp]
    \small
    \setlength\tabcolsep{2pt}
  \centering
  \caption{Standard and robust accuracy of different Adversarial Training (AT) and Adversarial Purification (AP) methods against PGD+EOT and AutoAttack $\ell_\infty (\epsilon = 8/255)$ on CIFAR-10. $\textbf{}^*$ utilizes half number of iterations for the attack due to the high computational cost. $\textbf{}^{\dag}$ indicates the requirement of extra data.}
    \label{linf-cifar10-28-10}
    \begin{tabular}{ccccc}
    \toprule
    \multirow{2}[0]{*}{Type } & \multirow{2}[0]{*}{Method}  & \multirow{2}[0]{*}{Standard Acc.} & \multicolumn{2}{c}{Robust Acc.} \\
          &       &                & PGD   & AutoAttack \\
    \midrule
    \rowcolor{gray!20} \multicolumn{5}{c}{WideResNet-28-10}\\
    \midrule
        &Gowal et al. \cite{improving}    & 88.54 & 65.93 & 63.38\\
    AT    &Gowal et al. \cite{uncovering}$^{\dag}$   & 87.51 & 66.01 & 62.76\\
        &Pang et al. \cite{pang2022robustness}
    & 88.62 & 64.95 &61.04\\
    \midrule
        & Yoon et al. \cite{score-based} & \cellcolor[rgb]{ .827,  .898,  .961}85.66±0.51 & \cellcolor[rgb]{ .867,  .922,  .969}33.48±0.86 & \cellcolor[rgb]{ .863,  .922,  .969}59.53±0.87 \\
        & Nie et al.  \cite{diffpure} & \cellcolor[rgb]{ .698,  .82,  .925}90.07±0.97 & \cellcolor[rgb]{ .635,  .78,  .91}56.84±0.59 & \cellcolor[rgb]{ .816,  .89,  .957}63.6±0.81 \\
        & Lee et al.  \cite{robust-evaluation} & \cellcolor[rgb]{ .698,  .816,  .925}90.16±0.64 & \cellcolor[rgb]{ .647,  .784,  .914}55.82±0.59 & \cellcolor[rgb]{ .729,  .839,  .933}70.47±1.53 \\
    AP    & Bai et al. \cite{Contrastive} & \cellcolor[rgb]{ .659,  .796,  .918}91.41 & \cellcolor[rgb]{ .71,  .827,  .929}49.22$^{\ast}$ & \cellcolor[rgb]{ .651,  .788,  .914}77.08 \\
        & Geigh et al. \cite{lorid}  & \cellcolor[rgb]{ .867,  .922,  .969}84.20 &   -    & \cellcolor[rgb]{ .867,  .922,  .969}59.14 \\
        & Lin et al. \cite{ATOP} & \cellcolor[rgb]{ .682,  .808,  .922}90.62 &   -    & \cellcolor[rgb]{ .702,  .82,  .929}72.85 \\
        & Ours  & \cellcolor[rgb]{ .608,  .761,  .902}\textbf{93.16±0.21} & \cellcolor[rgb]{ .608,  .761,  .902}\textbf{59.38±1.25} & \cellcolor[rgb]{ .608,  .761,  .902}\textbf{80.45±1.54} \\

    \bottomrule
    \end{tabular}%
\end{table}%

\begin{table}[htbp]
    \small
    \setlength\tabcolsep{2pt}
  \centering
  \caption{Standard and robust accuracy of different Adversarial Training (AT) and Adversarial Purification (AP) methods against PGD+EOT and AutoAttack $\ell_\infty (\epsilon = 8/255)$ on CIFAR-10.  $\textbf{}^*$ The number of iterations for the attack is half that of the other methods for less computational overhead. $\textbf{}^{\dag}$ indicates the requirement of extra data.}
    \label{linf-cifar10-70-16}
    \begin{tabular}{ccccc}

    \toprule
    \multirow{2}[0]{*}{Type } & \multirow{2}[0]{*}{Method}  & \multirow{2}[0]{*}{Standard Acc.} & \multicolumn{2}{c}{Robust Acc.} \\
          &       &                & PGD   & AutoAttack \\
    \midrule
     \rowcolor{gray!20} \multicolumn{5}{c}{WideResNet-70-16}\\
    \midrule
        &Rebuffi et al. \cite{rebuffi2021fixing}$^{\dag}$   & 92.22 & 69.97 &66.56\\
    AT    &Gowal et al. \cite{gowal2020uncovering}$^{\dag}$    & 91.10 & 68.66 &66.10\\
        &Gowal et al. \cite{gowal2021improving}   & 88.75 & 69.03 &65.87\\
    \midrule

        & Yoon et al. \cite{score-based} & \cellcolor[rgb]{ .804,  .882,  .953}86.76±1.15 & \cellcolor[rgb]{ .867,  .922,  .969}37.11±1.35 & \cellcolor[rgb]{ .851,  .914,  .965}60.86±0.56 \\
        & Nie et al.  \cite{diffpure} & \cellcolor[rgb]{ .698,  .816,  .925}90.43±0.60 & \cellcolor[rgb]{ .71,  .824,  .929}51.13±0.87 & \cellcolor[rgb]{ .8,  .882,  .953}66.06±1.17 \\
        & Lee et al. \cite{robust-evaluation} & \cellcolor[rgb]{ .694,  .816,  .925}90.53±0.1 & \cellcolor[rgb]{ .643,  .784,  .914}56.88±1.06 & \cellcolor[rgb]{ .757,  .855,  .941}70.31±0.62 \\
        & Bai et al.  \cite{Contrastive} & \cellcolor[rgb]{ .62,  .769,  .906}92.97 & \cellcolor[rgb]{ .733,  .839,  .937}48.83$^{\ast}$ & \cellcolor[rgb]{ .667,  .8,  .918}79.10 \\
    AP    & Geigh et al.  \cite{lorid} & \cellcolor[rgb]{ .867,  .922,  .969}84.60 &   -    & \cellcolor[rgb]{ .796,  .878,  .953}66.40 \\
        & Geigh et al. \cite{lorid} & \cellcolor[rgb]{ .8,  .882,  .953}86.90 &   -    & \cellcolor[rgb]{ .867,  .922,  .969}59.20 \\
        & Lin et al. \cite{ATOP} & \cellcolor[rgb]{ .651,  .788,  .914}91.99 &    -   & \cellcolor[rgb]{ .694,  .816,  .925}76.37 \\
        & Ours  & \cellcolor[rgb]{ .608,  .761,  .902}\textbf{93.36±0.59} & \cellcolor[rgb]{ .608,  .761,  .902}\textbf{59.79±0.61} & \cellcolor[rgb]{ .608,  .761,  .902}\textbf{84.83±0.59} \\

    \bottomrule
    \end{tabular}%
\end{table}%
\begin{table}[htbp]
    \small
    \setlength\tabcolsep{1pt}
  \centering
  \caption{Standard and robust accuracy against PGD+EOT and AutoAttack $\ell_2 (\epsilon = 0.5)$ on CIFAR-10. Adversarial Training (AT) and Adversarial Purification (AP) methods are evaluated. $\textbf{}^\ast$The number of iterations for the attack is half that of the other methods for less computational overhead. $\textbf{}^{\dag}$ indicates the requirement of extra data. $\textbf{}^\ddag$ adopts  WideResNet-34-10 as the backbone, with the same width but more layers than the default one.}
\label{l2-cifar10-28-10}  
    \begin{tabular}{ccccc}
    \toprule
    \multirow{2}[0]{*}{Type } & \multirow{2}[0]{*}{Method}  & \multirow{2}[0]{*}{Standard Acc.} & \multicolumn{2}{c}{Robust Acc.} \\
          &          & & PGD   & AutoAttack \\
    \midrule
    \rowcolor{gray!20} \multicolumn{5}{c}{WideResNet-28-10}\\
    \midrule
        &Rebuffi et al. \cite{rebuffi2021fixing}$^{\dag}$ & 91.79 & 85.05 &78.80 \\
    AT    &Augustin et al.  \cite{augustin2020adversarial}$^{\ddag}$ & 93.96 & 86.14&78.79 \\
        &Sehwag et al.  \cite{pang2022robustness}$^{\ddag}$ 
    & 90.93 & 83.75&77.24 \\
    \midrule
        & Yoon et al. \cite{score-based} & \cellcolor[rgb]{ .827,  .898,  .961}85.66±0.51 & \cellcolor[rgb]{ .867,  .922,  .969}73.32±0.76 & \cellcolor[rgb]{ .776,  .867,  .945}79.57±0.38 \\
        & Nie et al. \cite{diffpure} & \cellcolor[rgb]{ .675,  .804,  .922}91.41±1.00 & \cellcolor[rgb]{ .753,  .851,  .941}79.45±1.16 & \cellcolor[rgb]{ .741,  .847,  .937}81.7±0.84 \\
        & Lee et al \cite{robust-evaluation}  & \cellcolor[rgb]{ .706,  .824,  .929}90.16±0.64 & \cellcolor[rgb]{ .675,  .804,  .922}83.59±0.88 & \cellcolor[rgb]{ .667,  .8,  .918}86.48±0.38 \\
        & Bai et al. \cite{Contrastive}  & \cellcolor[rgb]{ .608,  .761,  .902}\textbf{93.75} & \cellcolor[rgb]{ .627,  .773,  .91}86.13$^{\ast}$ & \cellcolor[rgb]{ .753,  .851,  .941}80.92 \\
    AP    & Geigh et al.  \cite{lorid} & \cellcolor[rgb]{ .863,  .922,  .969}84.40 &   -    & \cellcolor[rgb]{ .8,  .882,  .953}77.90 \\
        & Geigh et al.  \cite{lorid} & \cellcolor[rgb]{ .867,  .922,  .969}84.20 &  -     & \cellcolor[rgb]{ .867,  .922,  .969}73.60 \\
        & Lin et al. \cite{ATOP} & \cellcolor[rgb]{ .694,  .816,  .925}90.62 &   -    & \cellcolor[rgb]{ .761,  .859,  .941}80.47 \\
       & Ours  & \cellcolor[rgb]{ .627,  .773,  .91}93.16±0.21 & \cellcolor[rgb]{ .608,  .761,  .902}\textbf{87.11±0.95} & \cellcolor[rgb]{ .608,  .761,  .902}\textbf{90.25±0.80} \\
    \bottomrule
    \end{tabular}%
\end{table}%

\begin{table}[htbp]
\small
\setlength\tabcolsep{1pt}
  \centering
  \caption{Standard and robust accuracy against PGD+EOT and AutoAttack $\ell_2 (\epsilon = 0.5)$ on CIFAR-10. Adversarial Training (AT) and Adversarial Purification (AP) methods are evaluated. ($\textbf{}^\ast$The number of iterations for the attack is half that of other methods for less computational overhead. $\textbf{}^{\dag}$ methods need extra data.)}
\label{l2-cifar10-70-16}  
    \begin{tabular}{ccccc}
    \toprule
    \multirow{2}[0]{*}{Type } & \multirow{2}[0]{*}{Method}  & \multirow{2}[0]{*}{Standard Acc.} & \multicolumn{2}{c}{Robust Acc.} \\
          &          & & PGD   & AutoAttack \\

    \midrule
    \rowcolor{gray!20} \multicolumn{5}{c}{WideResNet-70-16}\\
    \midrule
            &Rebuffi et al. \cite{rebuffi2021fixing}$^{\dag}$ & 95.74 & 89.62&82.32 \\
    AT    &Gowal et al. \cite{gowal2020uncovering}$^{\dag}$  & 94.74 & 88.18 & 80.53\\
        &Rebuffi et al. \cite{rebuffi2021fixing} & 92.41 & 86.24 & 80.42\\
    \midrule
        & Yoon et al. \cite{score-based} & \cellcolor[rgb]{ .867,  .922,  .969}86.76±1.15 & \cellcolor[rgb]{ .867,  .922,  .969}75.666±1.29 & \cellcolor[rgb]{ .867,  .922,  .969}80.43±0.42 \\
        & Nie et al. \cite{diffpure} & \cellcolor[rgb]{ .659,  .792,  .918}92.15±0.72 & \cellcolor[rgb]{ .71,  .827,  .929}82.97±1.38 & \cellcolor[rgb]{ .808,  .886,  .957}83.06±1.27 \\
        & Lee et al \cite{robust-evaluation}  & \cellcolor[rgb]{ .722,  .831,  .933}90.53±0.14 & \cellcolor[rgb]{ .694,  .816,  .925}83.75±0.99 & \cellcolor[rgb]{ .749,  .851,  .941}85.59±0.61 \\
    AP    & Bai et al. \cite{Contrastive}  & \cellcolor[rgb]{ .624,  .773,  .91}92.97 & \cellcolor[rgb]{ .682,  .808,  .922}84.37$^{\ast}$ & \cellcolor[rgb]{ .808,  .886,  .957}83.01 \\
        & Lin et al. \cite{ATOP} & \cellcolor[rgb]{ .663,  .796,  .918}91.99 &    -   & \cellcolor[rgb]{ .847,  .91,  .965}81.35 \\
        & Ours  & \cellcolor[rgb]{ .608,  .761,  .902}\textbf{93.36±0.59} & \cellcolor[rgb]{ .608,  .761,  .902}\textbf{87.7±0.97} & \cellcolor[rgb]{ .608,  .761,  .902}\textbf{91.71±1.09} \\
    \bottomrule
    \end{tabular}%
\end{table}%

\begin{table}[htbp]
\small
\setlength\tabcolsep{4pt} 
  \centering
  \caption{Standard and robust accuracy against BPDA+EOT $\ell_\infty(\epsilon = 8/255)$ on CIFAR-10.}
  \label{bpda}
    \begin{tabular}{ccccr}
    \toprule

    \multicolumn{2}{c}{\multirow{2}[0]{*}{Method}} & \multirow{2}[0]{*}{Purification} & \multicolumn{2}{c}{Accuracy} \\
    \multicolumn{2}{c}{} &       & Standard & \multicolumn{1}{c}{Robust} \\
    \midrule
     \rowcolor{gray!20} \multicolumn{5}{c}{WideResNet-28-10}\\
    \midrule
    \multicolumn{2}{c}{Song et al. \cite{pixeldefend}} & Gibbs Update & 95.00    & \multicolumn{1}{c}{9.00} \\
    \multicolumn{2}{c}{Yang et al. \cite{menet}} & Mask+Recon & 94.00    & \multicolumn{1}{c}{15.00} \\
    \multicolumn{2}{c}{Hill et al. \cite{hill2021stochastic}} & EBM+LD & 84.12 & \multicolumn{1}{c}{54.90} \\
    \midrule
    \multicolumn{2}{c}{Yoon et al. \cite{score-based}} & DSM+LD &\cellcolor[rgb]{ .867,  .922,  .969} 85.66±0.51 & \cellcolor[rgb]{ .867,  .922,  .969}66.91±1.75\\
    \multicolumn{2}{c}{Nie et al. \cite{diffpure}} & Diffusion &\cellcolor[rgb]{ .714,  .827,  .929} 90.07±0.97 &  \cellcolor[rgb]{ .643,  .784,  .914}81.45±1.51 \\
    \multicolumn{2}{c}{Bai et al. \cite{Contrastive}} & Diffusion &\cellcolor[rgb]{ .667,  .8,  .918} 91.37±1.21 &  \cellcolor[rgb]{ .643,  .784,  .914}85.51±0.81 \\
    \multicolumn{2}{c}{Wang et al. \cite{GDMAP}} & Diffusion &\cellcolor[rgb]{ .718,  .831,  .933} 89.96±0.40 &\cellcolor[rgb]{ .765,  .859,  .945} 75.59±1.26 \\
    \multicolumn{2}{c}{Song et al. \cite{mimic}} & Diffusion &\cellcolor[rgb]{ .722,  .831,  .933} 89.88±0.35 &\cellcolor[rgb]{ .608,  .761,  .902} \textbf{88.43±0.83} \\
    \multicolumn{2}{c}{Lee et al. \cite{robust-evaluation}} & Diffusion &\cellcolor[rgb]{ .71,  .827,  .929} 90.16±0.64 & \cellcolor[rgb]{ .612,  .765,  .906} 88.40±0.88 \\
    \multicolumn{2}{c}{Ours} & Diffusion & \cellcolor[rgb]{ .608,  .761,  .902}\textbf{ 93.04±0.65}&\cellcolor[rgb]{ .631,  .776,  .91} 86.72±0.20  \\
    \bottomrule
    \end{tabular}%
\end{table}%



\subsection{Experimental Results}
\label{Experimental Results}
\textbf{CIFAR-10.}\quad  We conducted comprehensive experiments on the CIFAR-10 dataset, evaluating our method against other approaches on two model architectures: WideResNet-28-10 and WideResNet-70-16. We assessed robust accuracy under three attack types: BPDA+EOT, PGD+EOT, and AutoAttack. We tested against the infinity norm attack, and as shown in Table \ref{linf-cifar10-28-10}, when the backbone is WideResNet-28-10, our method significantly outperforms other baselines in both standard accuracy and robust accuracy, exceeding by 3.00\% in standard accuracy and leading by 3.56\% and 3.37\% under PGD and AutoAttack, respectively. From Table \ref{linf-cifar10-70-16}, we see that when WideResNet-70-16 is used as the backbone, our method exceeds the state-of-the-art (SOTA) by 1.37\% in standard accuracy and leads by 2.91\% and 5.73\% under PGD and AutoAttack, respectively. We also evaluated the accuracy of our method and other baselines under the Euclidean norm attack. From Table \ref{l2-cifar10-28-10}, we observe that, with WideResNet-28-10 as the backbone, although our method lags behind \cite{Contrastive} by 0.59\% in standard accuracy, it leads by 0.98\% and 3.77\% in robust accuracy. Due to the significant time overhead of \cite{Contrastive}, which only iterates 100 times under PGD—half of what others do—our method outperforms more under PGD attacks. From Table \ref{l2-cifar10-70-16}, we see that when WideResNet-70-16 is the backbone, our method outperforms other baselines, leading by 1.21\% in standard accuracy and by 3.33\% and 6.12\% in robust accuracy under PGD and AutoAttack, respectively. Additionally, we applied the BPDA+EOT attack, which approximates differentiability. From Table \ref{bpda}, we observe that although robust accuracy lags behind \cite{mimic} by 1.71\%, our method still leads in standard accuracy by 3.16\%, resulting in an average advantage of 0.73\%. We also measured the distance between purified samples and the original samples in feature space, shown in the Appendix. A smaller distance indicates that the purified samples retain more consistent semantic information with the original samples (further discussion in the appendix). From Figure \ref{joint-distribution}, we find that the distribution of purified samples by our method is most similar to the original samples compared to other methods.
Overall, our method outperforms others, demonstrating its effectiveness in preserving semantic information while eliminating adversarial perturbations.

\begin{table}[htbp]
\small
\setlength\tabcolsep{4pt}
  \centering
  \caption{Standard and robust accuracy against PGD+EOT $\ell_\infty (\epsilon = 8/255)$ on ImageNet. ResNet-50 is used as the classifier.}
    \label{linf-imagenet}
    \begin{tabular}{cccc}
    \toprule
    \multirow{2}[0]{*}{Type} & \multirow{2}[0]{*}{Method} & \multicolumn{2}{c}{Accuracy} \\
          &       & Standard & Robust \\
    \midrule
    \rowcolor{gray!20} \multicolumn{4}{c}{ResNet-50}\\
    \midrule
        & Salman et al. \cite{transferbetter} & 63.86 & 39.11 \\
    AT    & Engstrom et al.  \cite{engstrom2019robustness}    & 62.42 & 33.20 \\
        & Wong et al. \cite{fastbetter} & 53.83 & 28.04 \\
    \midrule
        & Nie et al. ($t^*=0.2$)  \cite{diffpure} & \cellcolor[rgb]{ .62,  .769,  .906}73.96 & \cellcolor[rgb]{ .867,  .922,  .969}40.63 \\
        & Nie et al. ($t^*=0.3$)  \cite{diffpure} & \cellcolor[rgb]{ .643,  .784,  .914}72.85 & \cellcolor[rgb]{ .769,  .859,  .945}48.24 \\
        & Nie et al. ($t^*=0.4$)  \cite{diffpure} & \cellcolor[rgb]{ .867,  .922,  .969}61.71 & \cellcolor[rgb]{ .855,  .914,  .969}41.67 \\
        & Bai et al. \cite{Contrastive}  & \cellcolor[rgb]{ .694,  .816,  .925}70.41 & \cellcolor[rgb]{ .855,  .914,  .969}41.7 \\
    AP    & Lee et al.  \cite{robust-evaluation} & \cellcolor[rgb]{ .671,  .8,  .922}71.42 & \cellcolor[rgb]{ .788,  .875,  .949}46.59 \\
        & Song et al.  \cite{mimic} & \cellcolor[rgb]{ .859,  .918,  .969}62.25 & \cellcolor[rgb]{ .729,  .835,  .933}51.14 \\
        & Geigh et al. \cite{lorid} & \cellcolor[rgb]{ .62,  .769,  .906}73.98 & \cellcolor[rgb]{ .655,  .792,  .918}56.54 \\
        & Ours  & \cellcolor[rgb]{ .608,  .761,  .902}\textbf{74.51±1.07} & \cellcolor[rgb]{ .608,  .761,  .902}\textbf{60.06±1.66}  \\
    \bottomrule
    \end{tabular}%
\end{table}%

\begin{figure}[htbp]
    \centering
    \includegraphics[width=0.5\textwidth]{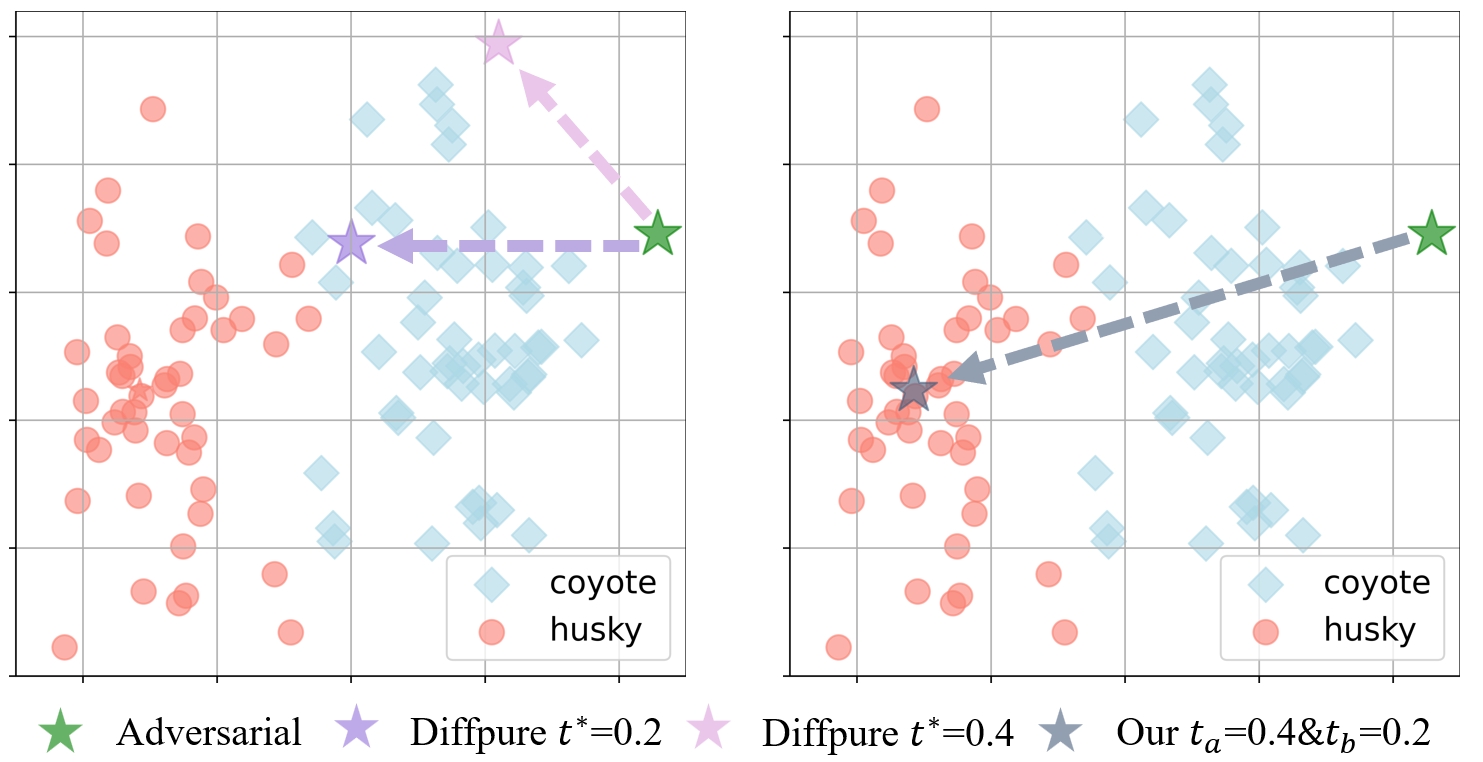}
    \caption{2D purification trajectories of Our method and Diffpure. For DiffPure (left), if $t^*$=0.2, although the purification direction is toward the original class, the adversarial perturbation cannot be completely eliminated; If $t^*$=0.4, the semantic information changes and the purification direction is no longer toward the original class. Our method (right) can eliminate adversarial perturbation and remain the semantic information.}
    \label{diffpure&our}
\end{figure}

\begin{figure}[!t]
    \centering
    \includegraphics[width=0.5\textwidth]{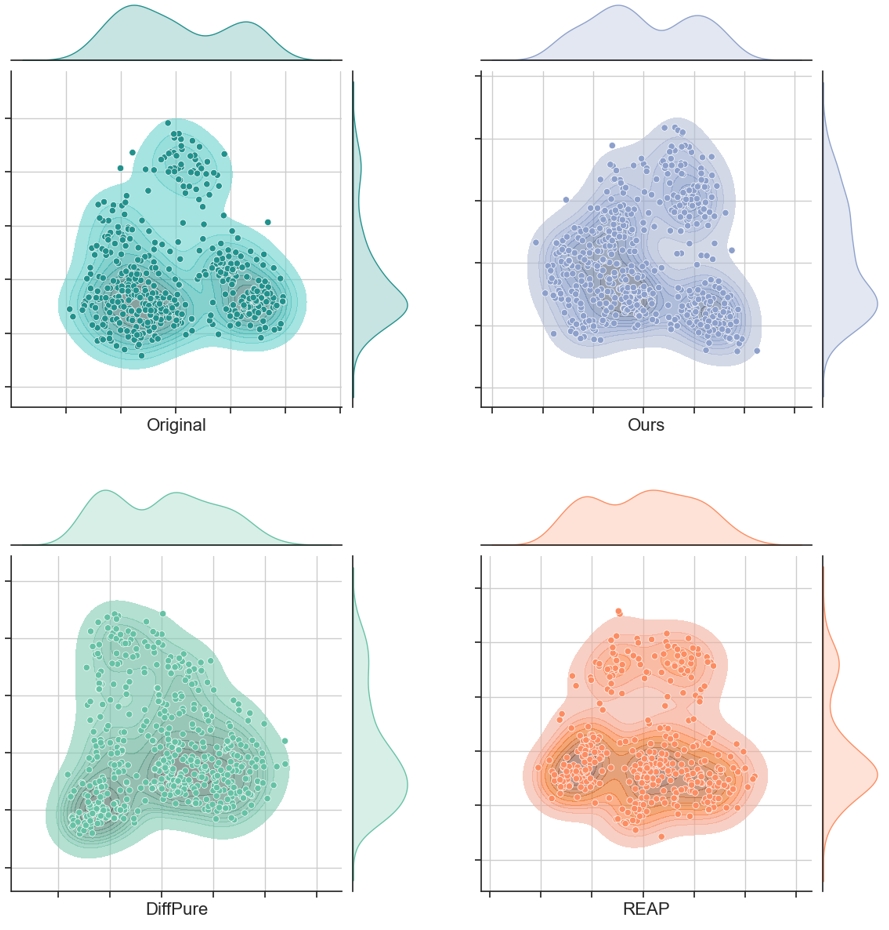}
    \caption{Joint distribution of the original samples and purified samples. The distributions of the purified samples by our method and the original samples are the most similar. }
    \label{joint-distribution}
\end{figure}

\textbf{ImageNet.}\quad We test ResNet50 as the backbone on the ImageNet dataset under PGD+EOT attacks, which is the same as \cite{diffpure,robust-evaluation,Contrastive}. From Table \ref{linf-imagenet}, our method leads competitors by 0.53\% in standard accuracy and by 3.52\% in robust accuracy, significantly surpassing the baseline. Additionally, we observe that the method from \cite{diffpure} achieves a standard accuracy of 73.96\% and a robust accuracy of 40.63\% when $t^* = 0.2$. When $t^* = 0.3$, standard accuracy drops to 72.85\% while robust accuracy rises to 48.24\%. However, at $t^* = 0.4$, both standard accuracy and robust accuracy decline to 61.71\% and 41.67\%, respectively.  Additionally, We visualize the 2D purification trajectories of our method and DiffPure using the T-SNE \cite{tsne} technique. From Figure \ref{diffpure&our}, we observe that when $t^* = 0.2$, the adversarial perturbation on the adversarial sample has not been fully purified, and it remains at the boundary between the two classes of samples. When $t^* = 0.4$, the semantic information of the adversarial example begins to change, and as a result, the purification trajectory no longer points toward the original classes. This indicates that applying the same noise intensity across an image cannot effectively balance semantic preservation and the elimination of adversarial perturbations. In contrast, our method adds varying noise intensities to different regions of an image, applying stronger noise to areas with higher classification weights to eliminate adversarial perturbations, while applying lighter noise to other regions to preserve semantic information. This approach allows our method to achieve good results in both robust accuracy and standard accuracy simultaneously. Our method achieves comparable standard accuracy to \cite{diffpure} at $t^* = 0.2$ but significantly outperforms it in robust accuracy, demonstrating the effectiveness of our approach.


\begin{figure}[htbp]
    \centering
    \includegraphics[width=0.5\textwidth]{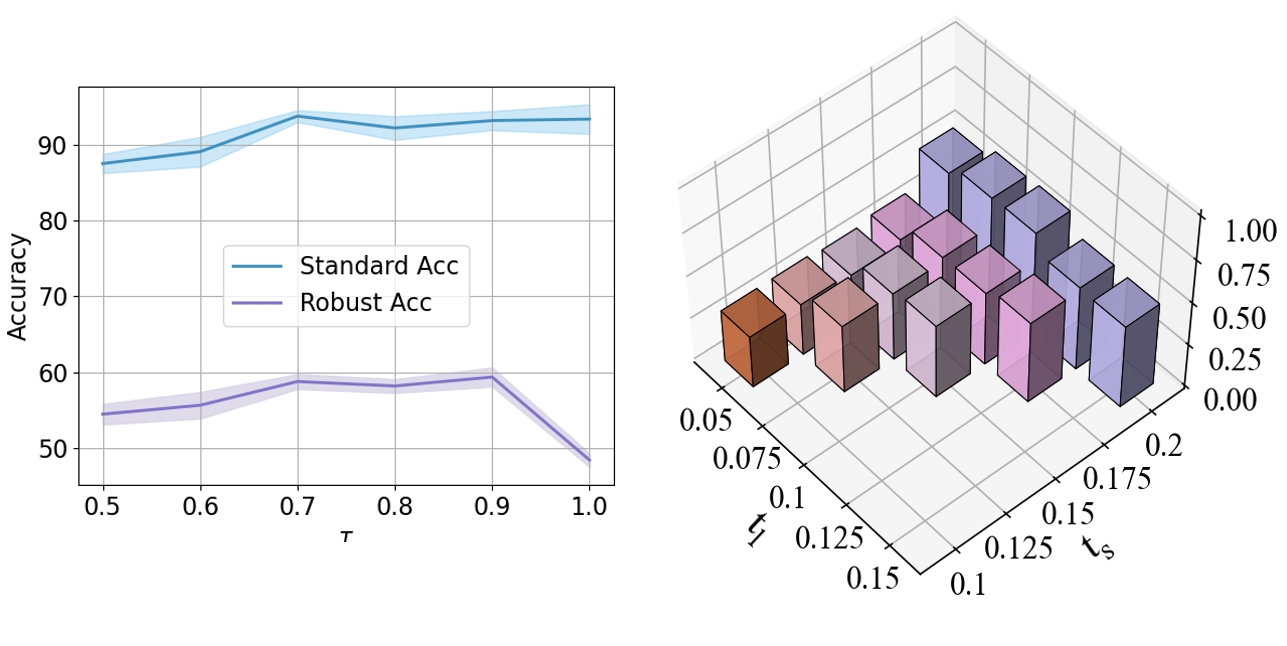}
    \caption{Robust and Standard Accuracy under different thresholds $\tau$  (left) and under different combinations of timesteps $t_l$ and $t_s$ (right).}
    \label{hyper}
\end{figure}
\subsection{Ablation Study}
\label{Ablation Study}
We first conducted a hyperparameter sensitivity analysis. Our method involves three hyperparameters, two of which are noises of different intensities, $t_l$ and $t_s$, and the third is a threshold, $\tau$, used to control the size of the attention mask. From Figure \ref{hyper}, we observe that as
$\tau$ increases (resulting in smaller mask areas), the Standard Accuracy also consistently increases. However, adversarial perturbations are preserved at this time, leading to poorer Robust Accuracy when 
$\tau$ is large. When $\tau$ is small, semantic information is damaged, which also results in weaker Robust Accuracy. Therefore, we choose $\tau$ to be around 0.8. Simultaneously, we can see that when \(t_l=0.2, t_s=0.05\), the Robust Accuracy is the highest. Here, \(t_s\) is less than the \(t^*\) used in  \cite{diffpure} to preserve more semantic information, while \(t_l\) is greater than \(t^*\) to eliminate adversarial perturbations. 
From Table \ref{linf-imagenet}, we can see that when we set $t_l=0.4$ and $t_s=0.2$, The Standard Accuracy is almost as high as the \(t^* = 0.2\) of  \cite{diffpure}, but at the same time, our Robust Accuracy significantly leads. This demonstrates the effectiveness of our method.\\
\begin{table}[htbp]
    \footnotesize
  \centering
  \caption{Standard Accuracy and Robust Accuracy for different block combinations. WideResNet28-10 servers as the backbone.}
    \begin{tabular}{ccccccc}
    \toprule
    \multicolumn{4}{c}{Encoder Block} & \multicolumn{3}{c}{Accuracy} \\
    E1    & E2    & E3    & E4    & Standard & Robust &Average \\
    \midrule
    \XSolidBrush & \Checkmark & \Checkmark & \Checkmark &92.77&58.00&75.38\\
     \Checkmark & \XSolidBrush & \Checkmark & \Checkmark &93.16&58.00&75.58\\
      \Checkmark & \Checkmark & \XSolidBrush & \Checkmark &93.35&57.61&75.48\\
       \Checkmark & \Checkmark & \Checkmark & \XSolidBrush &91.66&58.85&75.25\\
       \midrule
       \Checkmark & \Checkmark & \Checkmark & \Checkmark & 93.16 &59.38&76.27\\
       \bottomrule
    \end{tabular}%
  \label{encoder-ablation}%
\end{table}%
We also explore whether it is necessary for each block to participate in constructing the attention mask. Taking WideResNet28-10 as an example, it can be divided into four blocks. We extract the output of the final ReLU function of each block as the activation map. From Table \ref{encoder-ablation}, we can see that removing any block results in a performance loss. Therefore, every block should be utilized.

\section{Conclusion}

We propose a novel forward process that adds noise of varying intensities based on the classifier's attention to different regions, achieving a balance between preserving semantic information and eliminating adversarial perturbations. The denoising process is tailored to our method to remove different-intensity noise within a noisy image. By replacing the multi-step resampling with an improved single-step resampling, time cost and memory usage are reduced significantly. Extensive experiments on three datasets show that our method outperforms other methods by a promising improvement against various adversarial attacks.
{
    \small
    \bibliographystyle{ieeenat_fullname}
    \bibliography{main}
}
\clearpage
\setcounter{page}{1}
\maketitlesupplementary


\section{Space and Time Analysis}
\label{Space and Time}
\begin{figure}[htbp]
    \centering
    \includegraphics[width=0.45\textwidth]{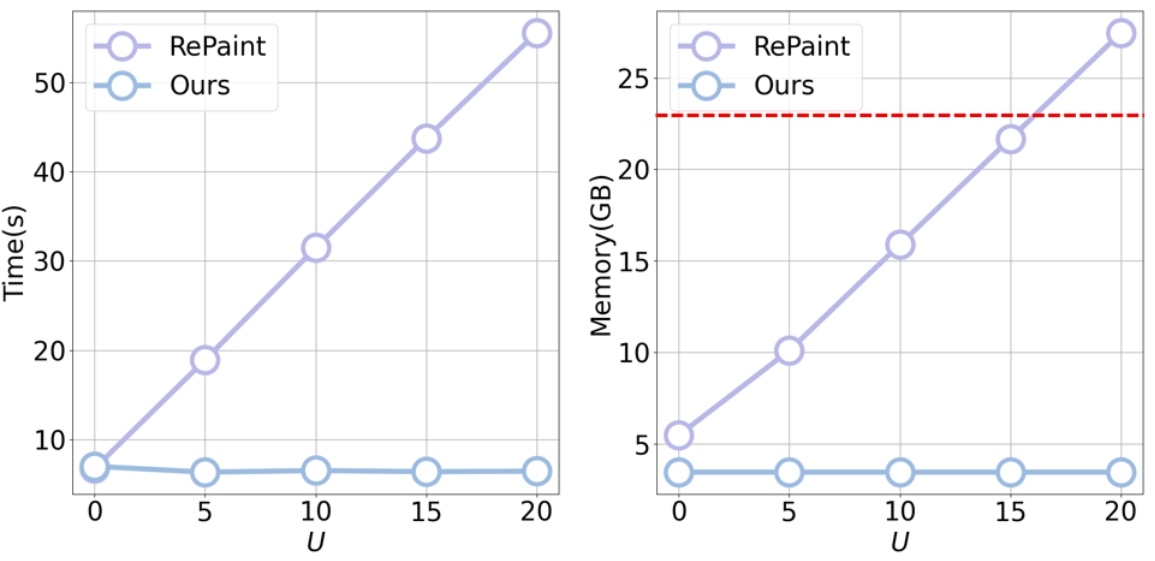}
    \caption{Time and space cost analysis and the dashed line represents the estimate. The red dashed line represents the maximum memory of consumer-grade GPUs. }
    \label{time&space}
\end{figure}

We compared the time cost and GPU memory usage before and after optimization. To evaluate the performance of our method against strong adaptive attacks, it is necessary to compute the gradients of our approach within the limited GPU memory. 
We selected an image from CIFAR-10 as a test, with a size of 32x32, and set the sampling steps to 10. From the Figure \ref{time&space}, we can find that, when we set $U=20$, our method takes 7 seconds to complete the sampling process, whereas the unoptimized method requires 64 seconds, reducing the time cost by 89\%. And for GPU memory consumption, our method reduced GPU memory usage from 27.4GB to 3.4GB, an 87\% reduction, enabling gradient computation on consumer-grade GPUs.

\section{Extra Experiments}

\begin{table}[htbp]
\setlength\tabcolsep{4pt}
  \centering
  \caption{Standard and robust accuracy against PGD+EOT $\ell_\infty (\epsilon = 8/255)$ on SVHN. Adversarial training methods
are evaluated on AutoAttack, and adversarial purification
methods are evaluated on PGD+EOT. WideResNet-28-10 is
used as a classifier except for Rade and Moosavi-Dezfooli, which uses ResNet-18.
}
    \label{linf-svhn}
    \begin{tabular}{cccc}
    \toprule
    \multirow{2}[0]{*}{Type} & \multirow{2}[0]{*}{Method} & \multicolumn{2}{c}{Accuracy} \\
          &       & Standard & Robust \\
    \midrule
    AT    &Rade and Moosavi-Dezfooli \cite{rade2022reducing} & 93.08 & 52.83 \\
    AT    & Gowal et al.   \cite{uncovering}    & 92.87 & 56.83 \\
    AT    &Gowal et al.  \cite{improving} & 94.15 & 60.90 \\
    \midrule
    AP    & Nie et al. ($t^*=0.2$)  \cite{diffpure}   & 92.57 &  35.54 \\
    AP    & Nie et al. ($t^*=0.3$)  \cite{diffpure}  & 85.74 & 45.12 \\
    AP    & Nie et al. ($t^*=0.4$)  \cite{diffpure}   & 73.63 & 43.55 \\

    AP    & Lee et al.  \cite{robust-evaluation} & \textbf{95.55} & 49.65 \\
    AP    & Ours &90.04& \textbf{62.50}\\
    \bottomrule
    \end{tabular}

\end{table}%
\textbf{SVHN}\quad We train on the SVHN dataset to obtain the unconditional SVHN checkpoint and the classifier weights for WideResNet-28-10, following the same experimental setup as described in \cite{robust-evaluation}. We also tested our method, along with other adversarial purification and adversarial training methods, on the SVHN dataset. We used the same Unet network architecture and parameters as in \cite{robust-evaluation} to train and obtain the pretrained weights on the SVHN dataset. The classifier model architecture used is WideResNet-28-10 and  robust accuracy is evaluated under the attack of PGD+EOT. From Figure \ref{linf-svhn} we can observe that although our method lags behind \cite{robust-evaluation} by 5.51\% in Standard Accuracy, it significantly outperforms by 12.85\% in Robust Accuracy. This demonstrates the effectiveness of our method. Additionally, we can observe that the method from  \cite{diffpure} achieves a Standard Accuracy of 92.57\% and a Robust Accuracy of 35.54\% when $t^*=0.2$. When $t^*=0.3$, Standard Accuracy drops to 85.74\% while Robust Accuracy rises to 45.12\%. However, at $t^*=0.4$, both Standard Accuracy and Robust Accuracy decline to 73.63\% and 43.55\%, respectively. This is consistent with the experimental results on the ImageNet dataset, indicating that globally adding noise to the image cannot balance both eliminating adversarial perturbations and preserving semantic information. At the same time, this further demonstrates the effectiveness of our method.

\section{Algorithm}

\begin{algorithm}[H]
\caption{Attention Mask Construction}
\label{mask}
\begin{algorithmic}
\REQUIRE Sample $x$, Value $\tau$, Classifier $F$ constitutes of  $m$ Bolcks with the activation output $A_{m}$.
\STATE $out = F(x)$
\COMMENT{Forward Propagation}
\STATE $\text{AM}_1 = \phi(\text{Bi}(\left( \sum_{i=1}^{C_m} |\boldsymbol{A}_{mi}|^p \right)^{\frac{1}{p}})$
\STATE $\text{AM}_2 = \phi(\text{Bi}(\left( \sum_{i=1}^{C_m} |\boldsymbol{A}_{mi}|^p \right)^{\frac{1}{p}})$
\STATE $\text{AM}_N = \phi(\text{Bi}(\left( \sum_{i=1}^{C_m} |\boldsymbol{A}_{mi}|^p \right)^{\frac{1}{p}})$
\STATE $    \mathcal{M} = \bigcup_{m=1}^M \mathbb{I}\ [\text{AM}_m>\tau]$
\RETURN $\mathcal{M}$
\end{algorithmic}
\end{algorithm}

\begin{algorithm}[H]
\caption{Adversarial Purification}
\label{AP}
\begin{algorithmic}
\REQUIRE Sample $x$, Classifier $F$, Number of Ensemble Runs $S$
\FOR{$s=1\  \textbf{to}\  S$}
\STATE Construct Attention Mask $\mathcal{M}$ using Algorithm \ref{mask}
\STATE Get Purified Image $x^{(s)}$ using \ref{sampling-argotithm} with $\mathcal{M}$
\ENDFOR
\STATE $\hat{y} = \arg\max_k \frac{1}{S} \sum_{s=1}^{S} \left[ F\left(x^{(s)}\right)\right]_k$

\RETURN $\hat{y}$
\end{algorithmic}
\end{algorithm}
\begin{figure}[H]
    \centering
    \includegraphics[width=0.45\textwidth]{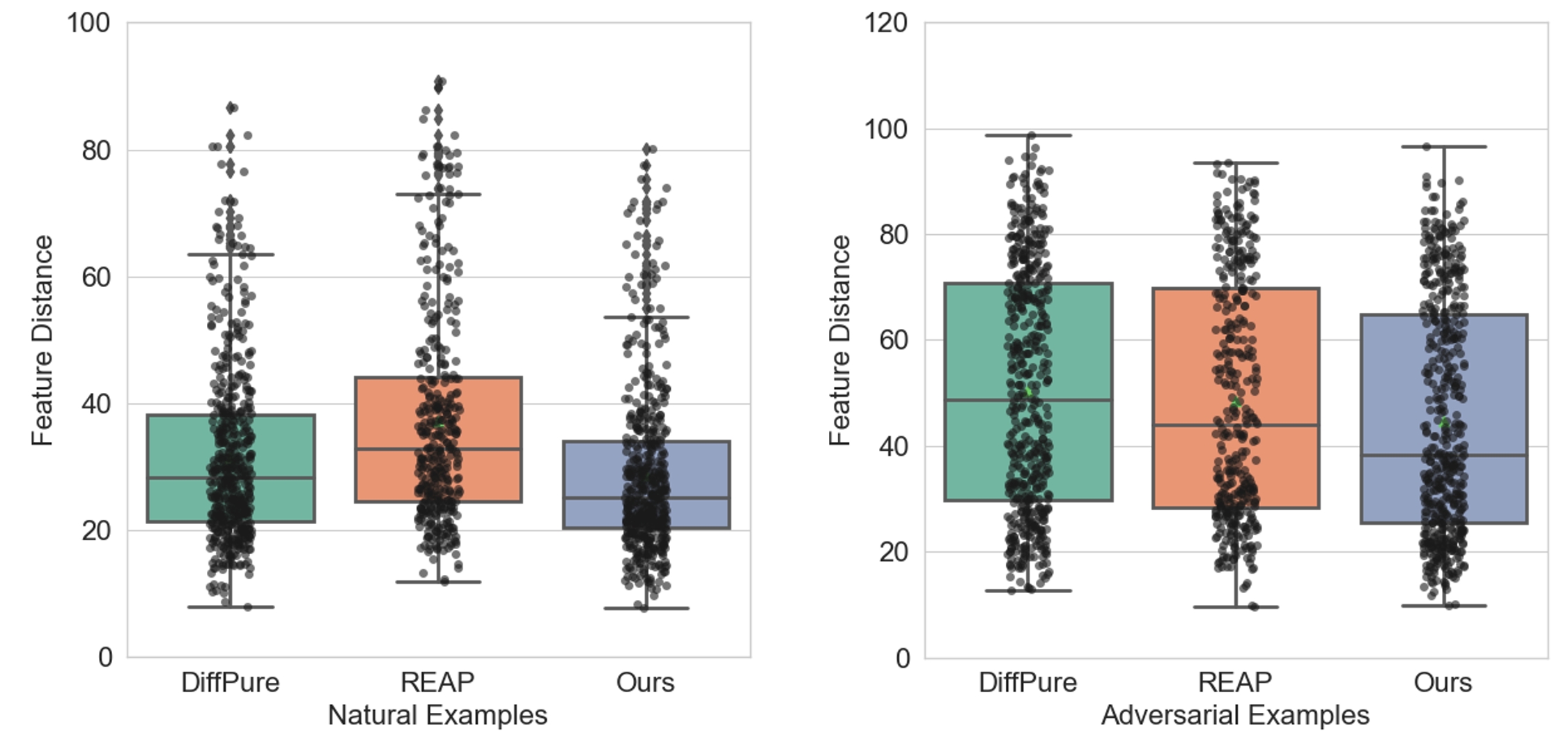}
    \caption{The distance between the purified samples and the original samples in the feature space. Our method can achieve the minimum feature distance on Natural Examples (left) and Adversarial Examples (right).}
    \label{feature-distance}
\end{figure}
\section{Discussion about Feature Distance}
To measure the degree of semantic consistency between the purified samples and the original samples, we need to select an appropriate metric. Intuitively, we could calculate the $L_p$ norm at the pixel level between the original samples and the purified samples. However, this approach presents a contradiction: for adversarial samples, although they may be close to the original samples at the pixel level, they can still lead the classifier to make incorrect classifications. Therefore, we do not choose pixel-level distance measurement. To also account for the changes in semantic information caused by adversarial perturbations, we select a distance measurement in the feature space of the classifier. If the semantics are similar, then they should also be close in feature space; if there are adversarial perturbations or semantic dissimilarities, then they should be far apart in the feature space which results in false predictions.





\end{document}